%% file: neurips_2026.tex
\documentclass{article}

    \PassOptionsToPackage{numbers, compress}{natbib}

\usepackage[preprint]{neurips_2026}

\usepackage[utf8]{inputenc} %
\usepackage[T1]{fontenc}    %
\usepackage{hyperref}       %
\usepackage{url}            %
\usepackage{booktabs}       %
\usepackage{amsfonts}       %
\usepackage{nicefrac}       %
\usepackage{microtype}      %
\usepackage{xcolor}         %
\usepackage{amsmath}

\title{Reasoning Arena: Trace Tournaments When Verifiable Rewards Fall Short}

\author{%
  Han Zhou\textsuperscript{1 2}\thanks{Work done as an AI Scientist Intern at Mistral AI.}\quad Adam X. Yang\textsuperscript{2} \quad Laurence Aitchison\textsuperscript{2} \quad Anna Korhonen\textsuperscript{1} \quad Albert Q. Jiang\textsuperscript{2}\\
  \textsuperscript{1}University of Cambridge \quad \textsuperscript{2}Mistral AI\\
  \texttt{hz416@cam.ac.uk} \quad \texttt{aj@mistral.ai}\\
}

\input{commands}

\begin{document}

\maketitle

\begin{abstract}
Reinforcement learning with verifiable rewards (RLVR) has become a leading paradigm for improving the reasoning ability of large language models through outcome-based supervision.
However, verifiable rewards frequently become uninformative at the group level: when all sampled traces of a given prompt receive identical rewards, group-relative advantage estimation provides no gradient signal, even though the traces may differ substantially in reasoning quality.
We propose \ours{}, an adaptive training framework that routes such non-diverse reward groups to a judge system instead of discarding them. Beyond examining the final answer, \ours{} constructs \emph{trace} tournaments, where reasoning traces are compared head-to-head to expose finer-grained preferences within the group, converting reasoning quality into rich relative reward signals.
To make reward estimation efficient, rather than exhaustively comparing every pair, each new trace is evaluated against a small, dynamically updated pool of previously generated traces as \emph{anchors} to efficiently establish a relative ranking. We then fit a Bradley-Terry model on the incomplete comparison graph, enabling scalable RL integration without quadratic pairwise comparisons.
Empirical results demonstrate that \ours{} consistently outperforms the RLVR baseline by 7.6\% on average in competition mathematics and coding benchmarks. By converting otherwise wasted zero-advantage samples into useful gradient updates, our method accelerates training by 27\% to 41\%, saving nearly 50\% of generation compute, and substantially improves overall reasoning performance. 

\end{abstract}

\section{Introduction}

Large language models (LLMs) have demonstrated remarkable progress in following instructions, generating coherent long-form responses, and conducting multi-step reasoning \citep{ouyang2022training,kojima2022large}. Recently, the reasoning capabilities of LLMs have been further advanced by the paradigm of reinforcement learning with verifiable rewards (RLVR), which trains LLMs directly on outcome-level signals provided by rule-based verifiers. Group-relative policy optimization methods, such as GRPO~\citep{shao2024deepseekmath}, drive this advancement by sampling a group of $N$ reasoning traces for a given prompt, scoring each trace with the verifier, and updating the policy using advantages computed relative to the group mean. This group-level contrast incentivizes the model to explore more effective thinking trajectories.

This paradigm is bottlenecked, however, by the lack of variance of rewards within each sampled group. Since GRPO relies on advantages computed relative to the group average, the estimator carries a gradient signal only when traces within a group receive different rewards. When the data difficulty is misaligned with the model's current capabilities, either too difficult or too easy, a substantial fraction of groups may end up with all traces marked incorrect or all correct. This results in a degenerate group where every advantage is exactly zero. We refer to such cases as \emph{non-diverse reward groups}. In these instances, the rewards collapse to a constant outcome across the entire group. Even though the underlying traces may differ substantially in reasoning quality, soundness, or conciseness, the verifier falls short by providing no signal to distinguish them. As a result, the group contributes zero gradient to policy updates, meaning the expensive compute used to generate these traces is completely wasted.

The existing literature has primarily approached the zero-variance problem through data curation, using prefiltering or difficulty prediction to avoid trivially easy or impossibly hard prompts \citep{yu2025dapo, zheng2026act}. However, discarding these non-diverse groups forfeits the fine-grained learning signals latent in their reasoning trajectories. A complementary line of work attempts to recover signals directly from these groups via entropy-guided advantage shaping \citep{le2026no, wu2025useless}. However, the reward reshaping relies entirely on the model's intrinsic token distribution and cannot distinguish between a rigorous proof and a confident hallucination. Consequently, the non-diverse reward group remains the fundamental barrier to learning meaningful reasoning behaviors on these prompts while making RL training inefficient.

To unlock the fine-grained learning signals hidden within non-diverse groups, we propose \ours{}, an adaptive training framework that establishes a hybrid-reward learning paradigm. 
\ours{} detects non-diverse groups online and dynamically routes them to a tournament-based reward system, while preserving the original verifiable reward for diverse groups. This ensures that verifiable rewards remain the gold standard whenever they produce informative advantages; the judge is only invoked precisely where the verifier falls short. For non-diverse reward groups, \ours{} constructs \emph{trace tournaments}: an LLM judge compares the reasoning \emph{traces} head-to-head within the group. Comparing intermediate traces rather than final answers is critical: it allows the judge to distinguish solutions that share identical verifier outcomes but differ in logical soundness and reasoning quality, eliciting fine-grained learning signals that have not been effectively utilized.
To make trace tournaments scalable in asynchronous RL, we introduce a live tournament strategy that pairs traces with live opponents, where we maintain a running pool of current best, worst, and median traces as anchors to efficiently establish a relative ranking. We then fit a Bradley-Terry (BT) model over the resulting incomplete comparison graph, enabling robust and efficient advantage estimation without incurring the high inference cost of full quadratic pairwise comparisons.

We evaluate \ours{} extensively across reasoning domains, including competition mathematics and code generation. Our empirical results demonstrate that by effectively extracting gradient signals from non-diverse reward groups, \ours{} consistently improves overall model performance over standard RLVR-only and pure LLM-as-a-judge baselines, preserving out-of-domain generalization. The gains are accompanied by faster training (27\% to 41\%) and a substantial reduction in generation compute of nearly 50\%, since non-diverse reward groups that would otherwise be discarded now provide meaningful gradients. Our cost analysis further shows that the live opponent selection strategy performs competitively with full round-robin tournament evaluations while effectively reducing the quadratic comparison cost to a linear scale.

Our contributions are summarized as follows: (1) We identify the inefficiency and wasted learning signals caused by discarded non-diverse groups in RLVR, and propose adaptive routing between verifiable and judge-based rewards as a principled remedy that preserves verifiable supervision wherever it is informative; (2) We propose \ours{} that introduces \emph{trace} tournaments as the reward mechanism for non-diverse reward groups, seamlessly scaled to RL training via a live tournament with Bradley-Terry estimation on incomplete graphs; (3) We demonstrate empirical gains on math and code reasoning tasks, along with improvements in training efficiency, OOD generalization, and the effective utilization of samples that would otherwise yield zero advantage.

\section{Related Work}
\sparagraph{Reinforcement Learning from Verifiable Rewards}
RLVR has emerged as the dominant paradigm for training reasoning models, moving from preference-based alignment~\citep{ouyang2022training,rafailov2023direct} to outcome-based supervision driven by rule-based verification~\citep{guo2025deepseek}. Group-relative policy optimization methods~\citep{shao2024deepseekmath, zheng2025group, chen2025minimax} rely heavily on the group-mean advantage estimator, consequently inheriting a critical zero-variance failure mode on non-diverse reward groups. DAPO~\citep{yu2025dapo} introduces dynamic sampling that filters out fully correct or fully incorrect groups during training, but this avoids rather than solves the problem: the generation compute is nevertheless wasted, and it forfeits the potential of learning from these samples. DEPO~\citep{tang2025towards} and GRESO~\citep{zheng2026act} similarly skip uninformative prompts at the data-curation or rollout-selection stage. A closer line of work derived signals from non-diverse groups themselves: RL-ZVP~\citep{le2026no} and ZAPO~\citep{wu2025useless} repurpose zero-variance prompts via entropy-guided advantage shaping, and RLPR~\citep{yu2025rlpr} replaces the verifier with a probabilistic surrogate. Since the recovered signal in all these methods is fundamentally intrinsic to the policy itself, it cannot distinguish a rigorous proof from a hallucinated derivation. \ours{} diverges by introducing an \emph{external} judge reward applied \emph{adaptively}. By composing verifiable and judge rewards at the per-group level within a single RLVR gradient step, non-diverse groups now gain rich gradient signals while the exact verifier continues to drive learning wherever it remains informative.

\rparagraph{LLM-as-a-Judge for Rich Rewards}
Beyond deterministic verifiers, LLM-as-a-judge has emerged as a flexible alternative for reward formulation. This approach was popularized by RLAIF~\citep{lee2024rlaif} and subsequently refined through pointwise rubric-based scoring~\citep{viswanathan2026checklists, gunjal2026rubrics, shao2025dr}. As pointwise judging is sensitive to prompt design and prone to surface-level bias~\citep{liu2026examining, zhou-etal-2024-fairer}, recent work shifts toward pairwise elicitation~\citep{liu2024aligning, rezaei2025online} and tournament-based ranking. Precisely, tournament rewards have been applied to non-verifiable domains~\citep{zhang2026arenarl, feng2026tourno}, including rubric-driven generation~\citep{jia2026open}, text-to-image~\citep{wang2025pref,wan2025maestro}, and even text-to-video generations~\citep{long2025vista}. They all target open-ended domains where no verifier exists, and use tournaments as the sole reward source applied uniformly to every group. \ours{} operates in a fundamentally different setting: reasoning tasks where a verifier exists but falls short of differentiating the intermediate reasoning trajectories of identical solution correctness. Rather than replacing the verifier, we embed tournaments \emph{within} RLVR and route rollouts to them adaptively. By invoking the judge strictly on non-diverse reward groups, \ours{} brings tournament-based judging into the verifiable-reward regime as a targeted complement, recovering the rich relative-ranking signal while preserving the exact verifier elsewhere.

\section{Preliminary}
\label{sec:preliminary}

\sparagraph{Reinforcement Learning from Verifiable Rewards}
We consider reinforcement learning for reasoning tasks with verifiable outcomes. Let $x \sim \mathcal{D}$ denote a prompt and $\pi_\theta$ a policy that generates a reasoning trace $y = (a_1, \dots, a_T)$. During training, a rollout engine samples a group of $N$ traces $\mathcal{G}(x) = \{y_i\}_{i=1}^{N}$ from the behavior policy $\pi_{\mathrm{old}}$. With a given binary verifiable reward function $R_v(x, y_i) = V(x, y_i) \in \{0, 1\}$, group-relative methods~\citep{shao2024deepseekmath} compute an advantage for each trace relative to the other samples within the same prompt group. We adopt CISPO~\citep{chen2025minimax} as our underlying RLVR algorithm in this work, which clips the importance-sampling weight rather than the policy-gradient term to retain learning signals from rare tokens. The RLVR objective is:
\begin{equation}
    \mathcal{J}_{\mathrm{RLVR}}(\theta) = \mathbb{E}_{x \sim \mathcal{D},\, \mathcal{G} \sim \pi_{\mathrm{old}}}\!\left[ \frac{1}{\sum_{i=1}^{N} |y_i|} \sum_{i=1}^{N} \sum_{t=1}^{|y_i|} \mathrm{sg}\!\bigl(\hat\rho_{i,t}(\theta)\bigr) \cdot A_i \cdot \log \pi_\theta(a_{i,t} \mid x, y_{i,<t}) \right],
    \label{eq:cispo}
\end{equation}
where $\mathrm{sg}(\cdot)$ denotes stop-gradient. The term $\hat\rho_{i,t}(\theta)$ is the clipped token-level importance-sampling weight, and $A_i$ is the group-relative trace advantage that scales every token-level update for trace $i$:
\begin{equation}
    \hat\rho_{i,t}(\theta) = \mathrm{clip}\!\left( \frac{\pi_\theta(a_{i,t} \mid x, y_{i,<t})}{\pi_{\mathrm{old}}(a_{i,t} \mid x, y_{i,<t})},\, 1 - \epsilon^{\mathrm{IS}}_{\mathrm{lo}},\, 1 + \epsilon^{\mathrm{IS}}_{\mathrm{hi}} \right),
    \qquad
    A_i = \frac{R_v(x, y_i) - \mu_{\mathcal{G}}}{\sigma_{\mathcal{G}} + \epsilon},
    \label{eq:rho-advantage}
\end{equation}
with $\mu_{\mathcal{G}} = \frac{1}{N} \sum_{j=1}^{N} R_v(x, y_j)$ and $\sigma_{\mathcal{G}}^{2} = \frac{1}{N} \sum_{j=1}^{N} ( R_v(x, y_j) - \mu_{\mathcal{G}} )^2$ denoting the within-group reward mean and variance, respectively. The defining characteristic of Eq.~\eqref{eq:rho-advantage} is that there must be a non-zero advantage for at least one sample in a group to produce a learning signal. Without a non-zero advantage function, the group cannot produce any meaningful gradient contribution.

\begin{wrapfigure}{r}{0.48\textwidth}
    \centering
    \vspace{-11.6mm}
\includegraphics[width=\linewidth]{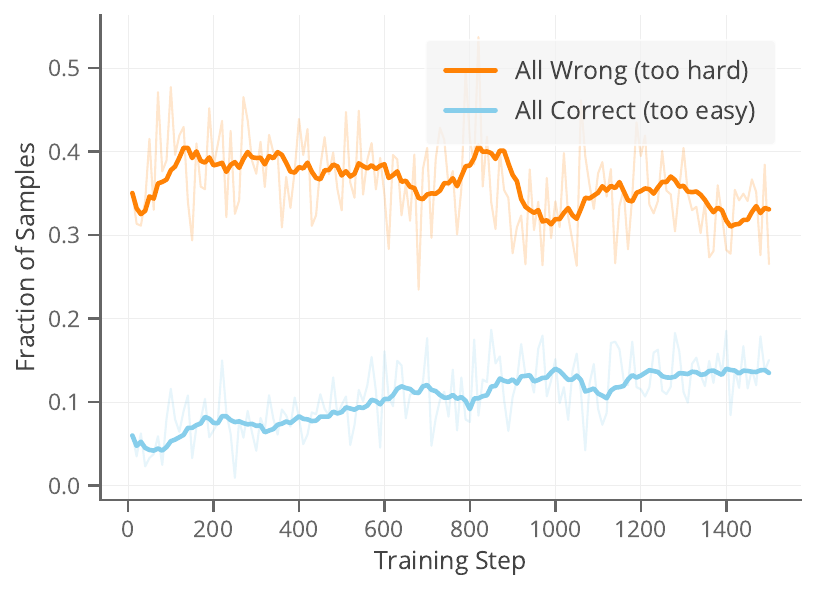}
    \vspace{-6.8mm}
    \caption{The presence of non-diverse reward groups is prevalent in RLVR. The fraction of non-diverse groups is decomposed into all-incorrect and all-correct groups for training with \texttt{Ministral-3-8B-Instruct}. All-wrong groups dominate early training, while all-correct groups increase as the policy improves; both regimes receive zero group-relative advantage under standard RLVR despite having already consumed rollout compute.}
    \label{fig:non_diverse_curve}
    \vspace{-3mm}
\end{wrapfigure}

\sparagraph{The Non-Diverse Reward Group Problem}
We formalize the reward diversity of a rollout group directly through its within-group reward variance:
\begin{equation}
    D(\mathcal{G}) = \mathbb{I}\!\left[\, \sigma_{\mathcal{G}} > 0 \,\right].
    \label{eq:diversity}
\end{equation}
A group is \emph{reward-diverse} when $D(\mathcal{G}) = 1$ and \emph{non-diverse} when $D(\mathcal{G}) = 0$. In the context of binary verifiable rewards, non-diverse groups manifest in exactly two forms: \emph{all-correct} groups, where $R_v(x, y_i) = 1$ for every trace, and \emph{all-incorrect} groups, where $R_v(x, y_i) = 0$ for every trace. By construction, this lack of reward diversity enforces a strict mathematical collapse:
\begin{equation}
    D(\mathcal{G}) = 0 \quad \Longleftrightarrow \quad \sigma_{\mathcal{G}} = 0 \quad \Longleftrightarrow \quad A_i = 0 \;\; \forall\, i \in \{1, \dots, N\}.
    \label{eq:zero-advantage}
\end{equation}
Consequently, a non-diverse group contributes absolutely no reward-driven policy gradient under Eq.~\eqref{eq:cispo}, regardless of how much the underlying trajectories differ in reasoning steps. This failure mode dominates two distinct training regimes: \emph{all-incorrect} groups are highly prevalent early in training when the policy remains weak, and \emph{all-correct} groups, which grow as the policy improves (Figure~\ref{fig:non_diverse_curve}). In modern asynchronous RL systems, the batch must still be filled to the target size before a training step can proceed: once a non-diverse group is derived, all of its $N$ already-generated traces are discarded. The engine must then sample additional rollouts to fill the batch, which is both costly and inefficient, as the per-trace generation cost dictates the overall pipeline latency. This structural inefficiency motivates our approach to extract and utilize the reasoning variation within these non-diverse groups, converting otherwise wasted compute into constructive gradient updates.

\section{Reasoning Arena}
\label{sec:method}

To address the non-diverse reward group problem, \ours{} adaptively routes each rollout group to distinct reward sources based on the variance of the received rewards. The primary motivation is to retain the exactness of the verifiable reward where it provides meaningful gradient signals, while dynamically falling back to a judge system precisely when the verifier falls short. For reward-diverse groups, we maintain the standard RLVR update; conversely, non-diverse groups are routed to a \emph{trace tournament}. Rather than evaluating final answers, this tournament conducts trace-by-trace comparisons to elicit granular signals regarding the reasoning quality of each trajectory. To scale this mechanism for asynchronous RL, we further introduce a live opponent strategy combined with Bradley-Terry fitting, which significantly reduces the computational overhead.

\sparagraph{Adaptive group routing}
The rule-based verifier $R_v$ from Eq.~\eqref{eq:rho-advantage} is highly efficient and, whenever a group exhibits reward diversity, $D(\mathcal{G}) = 1$, it is capable of separating the generated traces to provide a robust gradient signal. By comparison, LLM judges are computationally expensive and inherently noisier. However, LLM-as-a-judge offers a distinct advantage: it can evaluate intermediate reasoning steps to rank trajectories that share the exact same final answer. This makes the judge perfectly suited for non-diverse reward groups, which is the precise scenario where the verifier falls short and yields zero gradient. To leverage the strengths of both approaches, \ours{} dynamically integrates these two reward mechanisms at the per-group level. It routes to a more informative reward source conditioned on whether the verifier successfully produces within-group variance:
\begin{equation}
R^\star(x,y_i)= \begin{cases} R_v(x,y_i), & \text{if } D(\mathcal{G})=1,\\ R_j(x,y_i), & \text{if } D(\mathcal{G})=0, \end{cases}\label{eq:routing}
\end{equation}
where $R_j$ denotes a judge-derived reward (e.g., pointwise scoring or trace tournaments) defined in Section~\ref{sec:method:reward}. Substituting this into the group statistics of Eq.~\eqref{eq:rho-advantage}, $R^\star$ yields the adaptive objective:
\begin{equation}
    \mathcal{J}_{\mathrm{RA}}(\theta) = \sum_{\mathcal{G}:D(\mathcal{G})=1} \mathcal{J}_{\mathrm{RLVR}}(\theta;\mathcal{G},R_v) + \sum_{\mathcal{G}:D(\mathcal{G})=0} \mathcal{J}_{\mathrm{Arena}}(\theta;\mathcal{G},R_j).\label{eq:adaptive-objective}
\end{equation}
On reward-diverse groups, Eq.~\eqref{eq:adaptive-objective} simply reduces to standard RLVR. The router isolates the judge's intervention strictly to groups that would otherwise contribute zero gradient. Therefore, the judge reward acts as a targeted regularization term that extracts hidden reasoning variations.

\sparagraph{From pointwise scoring to trace tournaments}
Since every trace in a non-diverse reward group shares an identical verifier outcome, the final answer provides no distinguishing information, forcing the judge to evaluate the underlying reasoning trajectory itself. A straightforward realization of $R_j$ is a pointwise judge evaluating each trace in isolation based on a predefined rubric, which we later evaluate as the \texttt{Adaptive Pointwise} baseline. However, absolute pointwise scores are notoriously difficult to calibrate consistently across different traces \citep{li-etal-2025-large-language-models, zhang2026arenarl}, and they are susceptible to superficial heuristics, such as length, formatting, and hedging \citep{zhou-etal-2024-fairer, liu2026examining}, which inadvertently incentivize reward hacking. Motivated by findings that pairwise preferences align more robustly with human judgment~\citep{liu2024aligning,zhou-etal-2024-fairer}, \ours{} employs pairwise comparisons between traces drawn from the same prompt and policy model. 

For any given pair of reasoning trace $(y_i, y_j)$, the judge processes the context $(x, y_i, y_j)$ and outputs a categorical verdict $v \in \{\mathrm{A}, \mathrm{B}, \mathrm{Tie}\}$. We map this verdict to a soft outcome for the first trace:
\begin{equation}
o_{ij} = \begin{cases} \gamma, & v=\mathrm{A},\\ 1/2, & v=\mathrm{Tie},\\ 1-\gamma, & v=\mathrm{B}, \end{cases} \qquad o_{ji}=1-o_{ij},\label{eq:soft-outcome}
\end{equation}
where $\gamma \in (1/2, 1]$ is a soft-margin hyperparameter that interpolates between an uninformative tie ($\gamma = 1/2$) and a deterministic win ($\gamma = 1$). The tie value is strictly constrained to $1/2$ by the antisymmetry condition $o_{ij} + o_{ji} = 1$, which is essential for ensuring unbiased win-rate aggregation and maintaining a valid probability distribution when fitting the Bradley-Terry model.

\textit{Order debiasing.}
Pairwise LLM judges often exhibit strong, prompt-dependent position bias~\citep{zhou2024batch}. To mitigate this, we adopt a simple permutation calibration \citep{wang-etal-2024-large-language-models-fair, zhou-etal-2024-fairer}. During sampling, the presentation order of every queried pair is randomized independently, converting systematic positional offsets into zero-mean noise. During aggregation, each observed match $(i, j, o_{ij})$ is symmetrically augmented with its mirror $(j, i, 1 - o_{ij})$, explicitly enforcing the antisymmetry $o_{ij} + o_{ji} = 1$ that a biased judge might otherwise violate.

\begin{figure}[!t]
    \centering
    \vspace{-5mm}
    \includegraphics[width=1.0\linewidth]{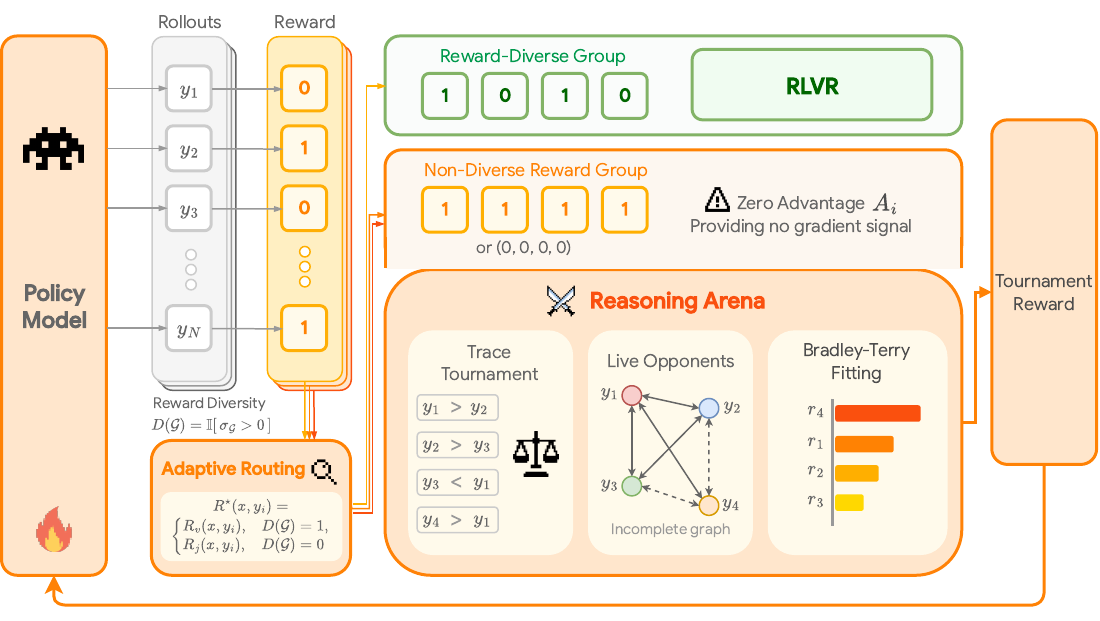}
    \vspace{-7mm}
    \caption{Overview of \ours{}. Rollout groups are adaptively routed based on their outcome diversity. Reward-diverse groups receive standard verifiable rewards. Conversely, non-diverse reward groups with zero advantage are dynamically routed to a trace tournament. By pairing traces with live opponents and fitting a Bradley-Terry model on the incomplete comparison graph, \ours{} efficiently extracts rich relative reward signals from reasoning traces.}
    \label{fig:method_overview}
\vspace{-1mm}
\end{figure}
\sparagraph{Deriving the tournament reward}
\label{sec:method:reward}
A naive tournament approach on a non-diverse group executes a full round-robin format, judging every pair $(y_i, y_j)$ to produce a complete match set of $\binom{N}{2}$ comparisons. The per-trace tournament reward then becomes its win-rate:
\begin{equation}
R_j^{\mathrm{wr}}(y_i) = \frac{1}{N-1}\sum_{j \ne i} o_{ij}.\label{eq:winrate}
\end{equation}
While mathematically unbiased on a complete graph, a round-robin incurs a prohibitive $O(N^2)$ inference cost per group, increasing the staleness in asynchronous RL pipelines under the same judge compute budget as the long-tail trace completes generation and gets paired to comparison late.

To resolve this, \ours{} executes a \emph{live} tournament that is fully parallelized with asynchronous RL. As soon as a new trace $y_k$ finishes generation, the live tournament dynamically selects three opponents as anchors for comparison from the traces that have already arrived. Specifically, the current best, worst, and median traces are chosen as the live opponents based on a live leaderboard ranked by $R_j^{\mathrm{wr}}(y_i)$ progressively. This setup includes extreme traces that always bracket the newly generated trace, while the median opponent acts as a middle anchor to rank the new generation. This strategy yields informative comparisons at a constant rate, reducing the per-group judging complexity from $O(N^2)$ to $O(N)$.

\sparagraph{Bradley-Terry Reward Aggregation}
Since the live opponent rule generates an incomplete, non-uniformly sampled comparison graph, utilizing the raw win-rate $R_j^{\mathrm{wr}}$ would confound the trace's true reasoning quality with the opponent-selection policy. 
To resolve this, we define the latent reasoning log-strength $\beta_i$ of each trace $y_i$ and estimate it using the Bradley-Terry (BT) model~\citep{bradley1952rank}. The probability that trace $y_i$ is preferred over $y_j$ by the judge is defined via the logistic sigmoid function:
\begin{equation}
    P(y_i \succ y_j) = \sigma(\beta_i - \beta_j) = \frac{1}{1 + \exp\bigl(-(\beta_i - \beta_j)\bigr)}.
    \label{eq:bt_prob}
\end{equation}
To robustly fit these strengths, we first construct a symmetrized match set $\widetilde{M}$ by mirroring every observed match: each outcome $(i, j, o_{ij})$ is explicitly duplicated as $(j, i, 1 - o_{ij})$. We then estimate the strength vector $\boldsymbol{\beta} \in \mathbb{R}^N$ by minimizing the L2-regularized soft cross-entropy loss:
\begin{equation}
    \mathcal{L}(\boldsymbol{\beta}) = -\sum_{(i,j,o_{ij}) \in \widetilde{M}} \left[ o_{ij} \log P(y_i \succ y_j) + (1-o_{ij}) \log P(y_j \succ y_i) \right] + \frac{1}{2} \|\boldsymbol{\beta}\|_2^2.
    \label{eq:bt_loss}
\end{equation}
This formulation natively supports the continuous soft outcomes $o_{ij} \in [0, 1]$ from Eq.~\eqref{eq:soft-outcome}, preserving the judge's relative rating granularity. The objective is also strictly convex, and we optimize it efficiently using the L-BFGS-B algorithm \citep{liu1989limited}. The L2 penalty ($\frac{1}{2}\|\boldsymbol{\beta}\|_2^2$) shrinks the ratings toward zero, preventing extreme values when the comparison graph is sparse and anchoring the global optimum even on disconnected subgraphs.
Finally, the fitted latent log-strengths $\boldsymbol{\beta}^\star$ are min-max normalized to produce the final scalar rewards $R_j \in [0, 1]$.

\section{Experiments}
\label{sec:experiments}
\sparagraph{Models and datasets}
We use \texttt{Ministral-3-8B-Instruct-2512}~\citep{liu2026ministral} as the policy model for follow-up RL training. For the primary LLM judge, we employ \texttt{DeepSeekMath-V2}~\citep{shao2025deepseekmath} to evaluate reasoning trajectories in the mathematical domain. To investigate the scalability of the reward source, we conduct additional ablations using \texttt{Qwen3-235B-A22B}~\citep{yang2025qwen3} and \texttt{Qwen3.5-122B-A10B} as the judge. Our RL training uses the STEM RL data mixture in \citep{liu2026ministral}. We explicitly filter out coding and visual-reasoning data during training, ensuring that the code task serves for out-of-distribution (OOD) evaluation. Our evaluation tasks include competition mathematics: AIME 2024/2025/2026, Beyond AIME~\citep{bytedance_seed_2025_beyondaime}, graduate-level domain reasoning: GPQA-Diamond~\citep{rein2024gpqa}, and code reasoning: LiveCodeBench v6~\citep{jain2025livecodebench}. We report the average pass ratio@16 for the math and GPQA-Diamond benchmarks, and pass@5 for LiveCodeBench to reduce variance.

\sparagraph{Baselines}
We compare \ours{} against representative baselines:
\begin{itemize}[leftmargin=5mm]
    \item \textbf{RLVR}~\citep{chen2025minimax}: We adopt CISPO as the underlying RLVR algorithm, and we directly conduct RL training on the policy model with verifiable rewards only, which serves as the primary baseline.
    \item \textbf{RLAIF}~\citep{lee2024rlaif}: We implement a pointwise LLM-as-a-judge baseline with the same policy optimization algorithm. The reward is generated via a DeepSeekMath-style proof-quality rubric, scoring each trace discretely in $\{0, 0.5, 1\}$ based on logical soundness.
    \item \textbf{ArenaRL}~\citep{zhang2026arenarl}: Learning from a tournament-only reward baseline. This is structurally equivalent to the round-robin tournament in \ours{} applied to all groups without adaptive routing, allowing us to ablate the impact of the routing mechanism.
\end{itemize}
In addition to the above baselines, we include \texttt{Adaptive Pointwise}, a \ours{} variant, where we replace the tournament reward with a pointwise rubric judge similar to RLAIF, while retaining the adaptive routing mechanism. In terms of our methods, we report the \ours{}, which routes non-diverse reward groups to a round-robin tournament using win-rate reward of Eq.~\eqref{eq:winrate}, and \ours{}-Live, which implements the live opponent selection rule in \S\ref{sec:method:reward} and Bradley-Terry reward aggregation.

\begin{table}[!t]
\centering
\caption{Main results on competition mathematics, scientific reasoning, and code benchmarks. Adaptive routing consistently improves over verifier-only RLVR and judge-only baselines. \ours{}-Live achieves the best average score, improving over RLVR by $+7.6\%$.}
\label{tab:main_results}
\resizebox{\textwidth}{!}{
\begin{tabular}{lccccccc}
\toprule
Method              & AIME 24 & AIME 25 & AIME 26 & Beyond AIME & GPQA-D & LCB v6 & Avg. \\
\midrule
RLVR  \citep{chen2025minimax}              & 58.5      & 43.8      & 46.0      & 28.2        & 54.8   & 46.7     & 46.3 \\
RLAIF \citep{lee2024rlaif}    & 53.1      & 46.9      & 48.3      & 27.6        & 56.9   & 50.7     & 47.3 \\
ArenaRL \citep{zhang2026arenarl}    & 60.4      & 50.2      & 56.9      & 31.9        & 59.6   & 50.4     & 51.6 \\
Adaptive Pointwise  & 62.9      & 51.5      & 54.4      & 31.9        & 58.5   & 48.5     & 51.3 \\
\midrule
\rowcolor{mistral-Beige-Light}\ours{}             & \textbf{66.7} & \textbf{54.8} & 54.6            & 33.6            & 59.5            & 51.6            & 53.5            \\
\rowcolor{mistral-pale-yellow}\ours{}-Live        & 63.5          & 51.7          & \textbf{59.0}   & \textbf{36.4}   & \textbf{60.5}   & \textbf{52.2}   & \textbf{53.9}   \\
\rowcolor{mistral-pale-yellow}$\Delta$(w.r.t. RLVR)  & \textcolor{plotgreen}{\textbf{+5.0}}& \textcolor{plotgreen}{\textbf{+7.9}}  & \textcolor{plotgreen}{\textbf{+12.9}} & \textcolor{plotgreen}{\textbf{+8.3}} & \textcolor{plotgreen}{\textbf{+5.7}} & \textcolor{plotgreen}{\textbf{+5.6}} & \textcolor{plotgreen}{\textbf{+7.6}}  \\
\bottomrule
\end{tabular}}
\end{table}
\begin{figure}[!t]
\vspace{-3mm}
\centering\includegraphics[width=1.0\linewidth]{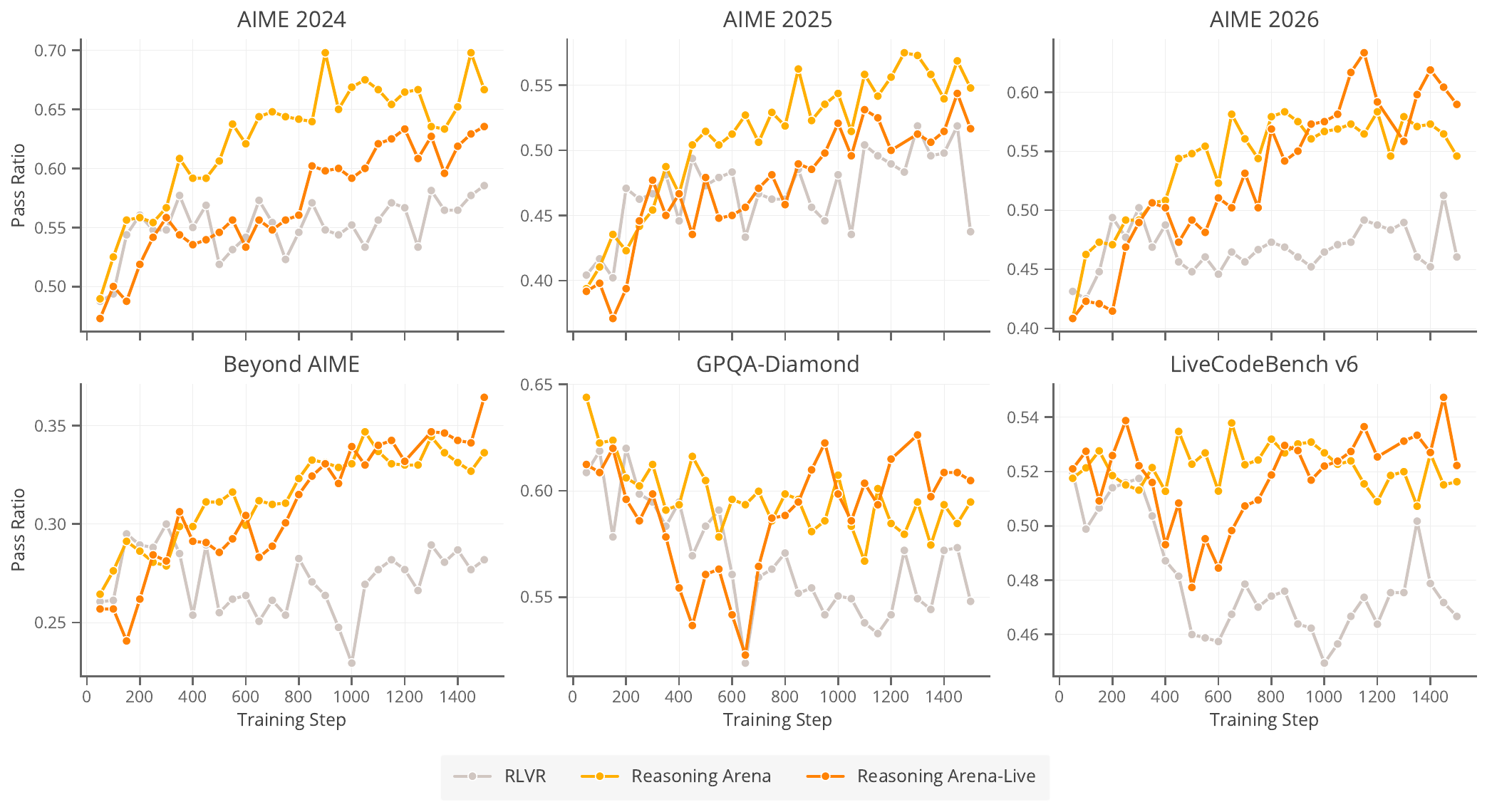}
\vspace{-6mm}
    \caption{Performance over RL training steps. \ours{} and \ours{}-Live improve steadily on in-domain math benchmarks and preserve gains on GPQA-Diamond and OOD LiveCodeBench, whereas verifier-only RLVR often plateaus or regresses after early training.}
    \label{fig:perfsteps}
\end{figure}

\sparagraph{Main results}
Table~\ref{tab:main_results} summarizes the final evaluation performance.
Standard RLVR reaches an average score of $46.3$, indicating that verifiable rewards alone are effective but leave substantial learning signal unused.
RLAIF improves the average only mildly, suggesting that replacing the verifier everywhere is not optimal in verifiable domains. 
In contrast, our adaptive routing approaches all give a consistent gain: \ours{}-Live reaches an average performance of $53.9$, outperforming RLVR by $7.6$ on average.
Notably, it improves AIME 2026 by $12.9$ points, with consistent improvements on the rest of tasks.
This demonstrates that converting hidden signals in non-diverse reward groups is critical for performance improvements, effectively expanding the usable learning signals while preserving the original verifiable supervision.

\begin{table}[t]
\centering
\small
\setlength{\tabcolsep}{4pt}
\vspace{-3mm}
\caption{Training efficiency and the judge reward costs with a group $N=8$ and the live tournament with three opponents. $\alpha$ is the fraction of groups that were routed to the trace tournament, so \emph{effective calls} are $\alpha \cdot c$. All metrics are averaged over the entire training. $^{*}$We note that the RLAIF step time is not directly comparable because that run generates substantially shorter responses, while longer responses increase wall-clock step time independently of generation efficiency.}
\label{tab:judge_cost}
\resizebox{\textwidth}{!}{
\begin{tabular}{lcccllc}
\toprule
Method & Judge calls $c$ & Routed groups $\alpha$ & Effective calls & Step time & \#Gens / step & Perf.\\
\midrule

RLVR \citep{chen2025minimax}               & $0$               & 0$\%$   & 0.0   & 17.7 & 908 & 46.3\\
RLAIF \citep{lee2024rlaif}              & $N=8$             & $100\%$ & $8.0$   & $4.2^{*}$ & $712$ & 47.3 \\
ArenaRL \citep{zhang2026arenarl}            & $\binom{N}{2}=28$ & $100\%$ & $28.0$  & 10.5 & $465$ & 51.6\\
Adaptive Pointwise  & $N=8$             & $47\%$  & $3.8$   & $10.6$ & $660$ & 51.3\\
\midrule
\rowcolor{mistral-Beige-Light}\ours{}             & $\binom{N}{2}=28$ & $45\%$  & $12.5$  & $10.4$ & 459 & 53.5\\
\rowcolor{mistral-pale-yellow}\ours{}-Live        & $18$              & 43$\%$  & 7.7   & 13.0 \textcolor{plotgreen}{\textbf{$\downarrow4.7$}} & 455 \textcolor{plotgreen}{\textbf{$\downarrow453$}} &53.9 \\
\bottomrule
\end{tabular}}

\end{table}

\begin{figure}[t]
\centering\includegraphics[width=1.0\linewidth]{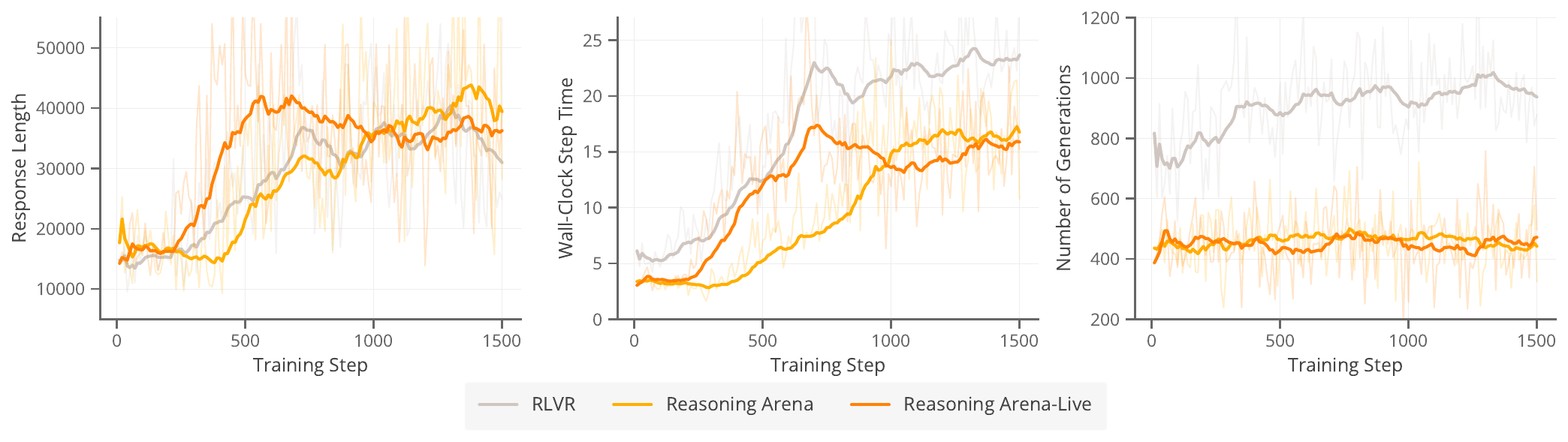}
\vspace{-6mm}
    \caption{Training efficiency and generation behavior. Though \ours{}-Live produces long reasoning traces, adaptive routing avoids repeatedly resampling zero-advantage groups, reducing \emph{half} of the number of generations needed per optimizer step. The live tournament keeps wall-clock step time below verifier-only RLVR while retaining the performance gains via tournament rewards.}
    \label{fig:efficiency}
\end{figure}

Figure~\ref{fig:perfsteps} compares performance over RL steps.
The margin between RLVR and \ours{} widens steadily during training, indicating that non-diverse reward groups provide usable gradients throughout optimization.
This divergence is particularly pronounced in later training stages. As the policy improves, all-correct groups become more prevalent. While standard group-relative RLVR assigns zero advantage to these groups, \ours{} successfully extracts fine-grained, trace-level preferences.
Furthermore, the sustained progress on LiveCodeBench and GPQA-Diamond demonstrates that our adaptive tournament preserves OOD generalization by learning from an external distribution, mitigating the domain overfitting frequently observed in pure verifier-driven training.

\sparagraph{Efficiency}
Table~\ref{tab:judge_cost} details the judge-call complexity for the evaluated reward configurations. With a group size of $N=8$, a full trace tournament dictates $\binom{8}{2}=28$ judge calls per routed group. The live opponent strategy reduces this to a maximum of $18$ $(1+2+3\times5)$ calls while also improving the overall performance average. Because the adaptive routing exclusively targets non-diverse groups, the empirical judge cost is heavily discounted by the fraction of reward-diverse groups present in the batch.
Crucially, this adaptive mechanism significantly improves overall training efficiency. In standard RLVR, a non-diverse group yields zero variance and therefore zero advantage, meaning the computational cost of generating those $N$ trajectories is entirely wasted. \ours{} instead reduces the wall-clock step time by 27\% to 41\% while saving nearly 50\% of the generations per step, massively reducing the generation overhead. By converting these previously discarded rollouts into rich tournament signals, \ours{} maximizes sample efficiency and training efficiency with a practically feasible judge costs for scalable RL training.

\begin{wrapfigure}{r}{0.48\textwidth}
    \centering
    \vspace{-4.5mm}
\includegraphics[width=\linewidth]{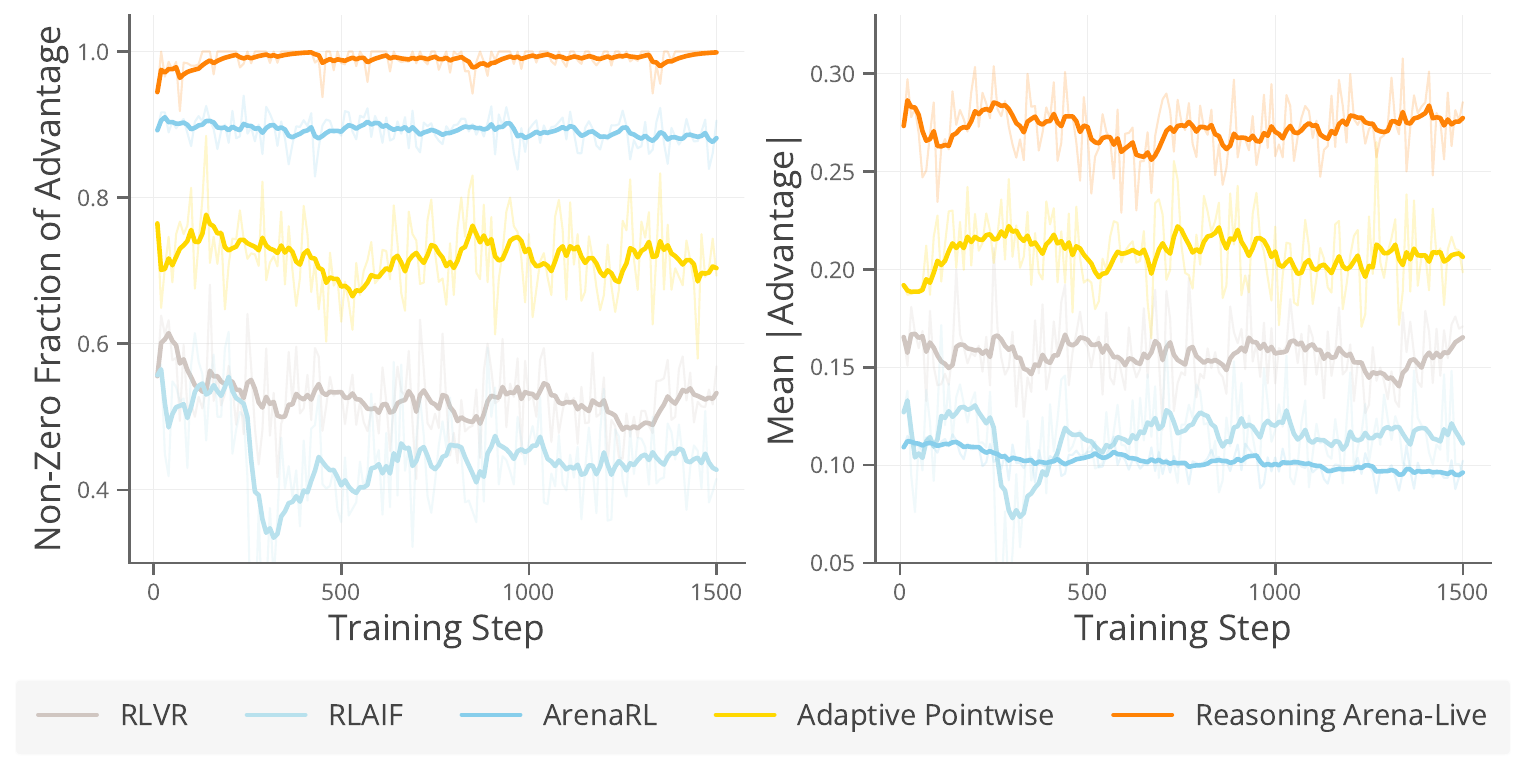}
    \vspace{-6mm}
    \caption{Tournament rewards restore usable advantages on non-diverse reward groups. Left: the fraction of traces with non-zero advantage remains highest for \ours{}-Live across training. Right: \ours{}-Live also yields the largest mean absolute advantage, indicating that trace tournaments provide denser and stronger credit assignment than verifier-only or pointwise judge rewards.}
    \label{fig:advantage}
    \vspace{-1mm}
\end{wrapfigure}
\sparagraph{Ablating the adaptive routing mechanism}
The comparison between ArenaRL and \ours{} highlights the effect of adaptive routing.
ArenaRL applies tournament rewards uniformly and improves over RLVR, but it discards the special reliability of exact verifiers on reward-diverse groups.
\ours{} instead uses tournaments only when the verifier provides no within-group contrast.
This strategic routing yields higher average performance than ArenaRL while concentrating computational resources on a targeted subset of the data. Figure \ref{fig:advantage} reflects the sample efficiency in RL by different learning strategies. It is notable that though ArenaRL comes with high non-zero fraction of advantage samples, the L1-norm is relatively small, whereas \ours{} effectively improves the sample utilization to nearly 100\% while providing denser reward signals. This observation further consolidates the adaptive routing mechanism in both improving training and sample efficiency.

\sparagraph{Tournaments vs. pointwise scoring}
The comparison of \ours{} against the \texttt{Adaptive Pointwise} ablates the form of the routed reward source.
Both methods employ the same adaptive routing, differing only in their scoring of non-diverse reward groups.
Trace tournaments improve the average from $51.3$ to $53.5$ with full tournaments.
This gap suggests that merely introducing an LLM judge is insufficient, but the form of reward evaluation is more critical. Pointwise scoring forces the judge to assign an absolute numerical value in isolation. This makes it notoriously difficult to calibrate across traces without exhaustive rubrics, leaving it vulnerable to superficial heuristics. Pairwise comparison, however, inherently grounds the evaluation by presenting the judge with two trajectories sharing identical verifiable outcomes. This naturally anchors the baseline, forcing the reward distribution to reflect true relative reasoning quality rather than formatting artifacts.

\sparagraph{Effectiveness of different judge models}
We further investigate the robustness of our approach with respect to various judge models. Replacing \texttt{DeepSeekMath-V2 (685B)} in \ours{} with broader models, such as \texttt{Qwen3-235B-A22B} and \texttt{Qwen3.5-122B-A10B}, reveals a scaling behavior. Higher-capacity models exhibit superior consistency in their pairwise preferences, particularly when judging complex, multi-step logical derivations. While \texttt{DeepSeekMath} is highly effective for mathematical domains, the \texttt{Qwen} variants also provide robust reward signals for non-diverse reward groups, showing that \ours{} is robust to judge models.
\begin{figure}
    \centering
    \vspace{-5mm}
    \includegraphics[width=1\linewidth]{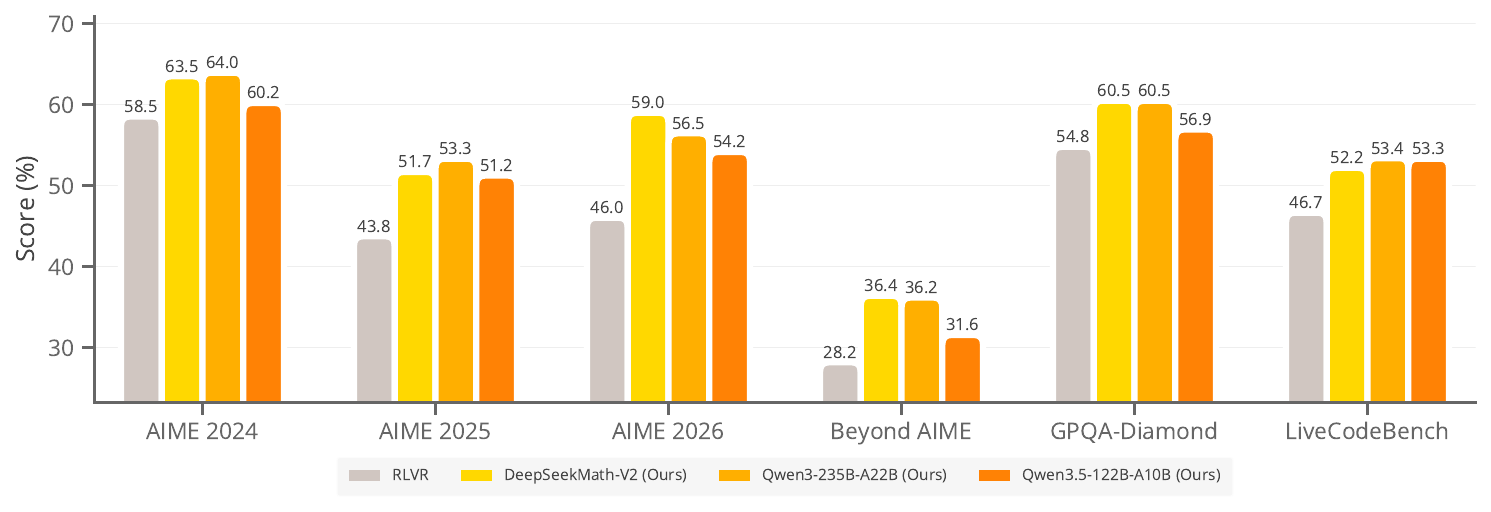}
    \vspace{-7mm}
    \caption{Extending the judge-model in \ours{}-Live to more model families. All judge choices substantially improve over RLVR, showing the robustness of \ours{}.}
    \label{fig:judge}
\end{figure}

\sparagraph{Analysis of the trace tournament mechanism}
To understand the qualitative advantage of trace tournaments, we analyzed the pairwise verdicts across both all-correct and all-incorrect groups (detailed in App.~\ref{sec:appendix:judge_ex2}). For all-correct groups, the judge consistently penalizes trajectories that arrive at the correct answer through logical gaps or incomplete justifications. Conversely, for all-incorrect groups, the tournament rescues partial learning signals that the exact verifier outright discards. When evaluating flawed responses, the judge systematically prefers structured problem-solving attempts, such as pattern derivation from small cases, over unsupported guesses. Consequently, the tournament provides dense and constructive feedback on the intermediate reasoning steps, incentivizing sound logic regardless of the final correctness.

\section{Conclusion}
\label{sec:conclusion}

We introduced \ours{}, an adaptive training framework designed to resolve the non-diverse reward group problem in RLVR. Standard group-relative methods fail to provide gradient signals when a deterministic verifier assigns identical outcomes to all traces in a sampled group. Rather than discarding these non-diverse reasoning traces, \ours{} adaptively routes them to a head-to-head trace tournament, leveraging an LLM judge to extract fine-grained relative preferences from the intermediate reasoning steps. To scale this mechanism for asynchronous RL training, we developed a live opponent selection strategy coupled with Bradley-Terry reward aggregation. This allows the framework to infer consistent rewards from sparse comparison graphs without the prohibitive cost of full quadratic evaluations. Empirical results across mathematical and code reasoning tasks demonstrate that \ours{} consistently outperforms standard RLVR and pointwise judge baselines. By converting otherwise wasted zero-advantage samples into informative gradient updates, \ours{} provides a principled, compute-efficient, and sample-efficient method for composing exact verification with LLM judges.

\section*{Acknowledgment}
We thank all other colleagues in Mistral AI for their valuable feedback. 
The work has been supported by the UK Research and Innovation (UKRI) Frontier Research Grant EP/Y031350/1 (the UK government’s funding guarantee for ERC Advanced Grants) awarded to Anna Korhonen at the University of Cambridge.

\bibliography{neurips_2026}
\bibliographystyle{neurips_2023}

\clearpage
\appendix

\section{Limitations and Future Work}
\label{app:limitation}
\ours{} is an adaptive reward framework orthogonal to specific model architectures and reinforcement learning algorithms. Extracting fine-grained signals from non-diverse reward groups, it yields substantial improvements over purely verifiable rewards. While our proposed adaptive routing effectively recovers these otherwise wasted signals, we note that the inherent prevalence of non-diverse reward groups remains dependent on the underlying data mixture and the base capabilities of the policy model. Nevertheless, \ours{} mitigates the computational burden of exhaustively prefiltering or re-grading the training corpus to construct a perfectly difficulty-matched dataset, which is itself costly and difficult to process per policy model to train.

Due to the strict latency and off-policy constraints inherent to asynchronous RL, passing complete reasoning trajectories can easily exceed 160,000 input tokens per pair. Consequently, \ours{} currently relies on the concise reasoning traces natively generated within the policy model's response trunk. In future work, intelligently incorporating truncated reasoning tokens could further enhance the judge's evaluation accuracy. Furthermore, the applicability of \ours{} could be naturally expanded to tool-use agents, where tens to hundreds of intermediate tool calls and high-level planning steps can serve as the reasoning trace for the judge to evaluate.

Although our approach significantly accelerates overall sample and training efficiency, it operates at the cost of additional GPU resources or API compute required to serve the judge models. This establishes a practical tradeoff among runtime acceleration, generation efficiency, and total resource allocation. Overall, \ours{} serves as a highly generalizable approach for overcoming the non-diverse reward group problem of verifiable rewards, offering a systematic framework for efficient RLVR optimization.

\section{Implementation Details}
\label{appendix:details}
\rparagraph{Models and Datasets}
In this work, we use the following models and datasets, all of which are released for research use under their respective licenses.

We use 1) \texttt{Ministral-3-8B-Instruct-2512}~\cite{liu2026ministral} (Mistral Research License: \url{https://mistral.ai/licenses/MRL-0.1.md}) as the policy model trained with online RL; 2) \texttt{DeepSeekMath-V2}~\cite{shao2025deepseekmath} (Apache 2.0 License: \url{https://huggingface.co/deepseek-ai/DeepSeek-Math-V2/blob/main/LICENSE}) as the LLM judge for pairwise tournament and pointwise scoring; 3) \texttt{Qwen3-235B-A22B}~\cite{yang2025qwen3} (Apache 2.0 License: \url{https://huggingface.co/Qwen/Qwen3-235B-A22B/blob/main/LICENSE}) and \texttt{Qwen3.5-122B-A10B} (Apache 2.0 License: \url{https://huggingface.co/Qwen/Qwen3.5-122B-A10B/blob/main/LICENSE}) as alternative judge backbones for ablation studies on judge families.

We evaluate on 1) AIME 2024, AIME 2025, and AIME 2026 (problems copyrighted by the Mathematical Association of America) as the competition-level mathematics benchmarks; 2) BeyondAIME~\cite{bytedance_seed_2025_beyondaime} (CC0 1.0 Universal Public Domain Dedication: \url{https://huggingface.co/datasets/ByteDance-Seed/BeyondAIME}); 3) GPQA-Diamond~\cite{rein2024gpqa} (CC-BY-4.0 License: \url{https://github.com/idavidrein/gpqa/blob/main/LICENSE}) as a graduate-level Google-proof Q\&A benchmark; and 4) LiveCodeBench~\cite{jain2025livecodebench} (MIT License: \url{https://github.com/LiveCodeBench/LiveCodeBench/blob/main/LICENSE}) as the coding benchmark.

\sparagraph{Setups}
We train Ministral 3 8B Instruct using FP8 quantization with a maximum sequence length of $81,920$, a constant learning rate at 4e-7 for $1500$ optimization steps with $50$ warm-up steps; each minibatch contains $480$ RL sampled rollouts, and the group size is $N=8$. We set $\epsilon^{\mathrm{IS}}_{\mathrm{hi}} = 2.0$ and $\epsilon^{\mathrm{IS}}_{\mathrm{lo}} = 1.0$ for one-sided clipping. A full round-robin tournament requires $\binom{8}{2} = 28$ judge inferences per non-diverse group, while \ours{}-Live with three opponents necessitates at most $18$ ($1+2+3*5$) inferences, such that early traces will only compare to the available opponents, and later traces compare against three anchors. The judge operates with a $32,768$-token context window and receives the ground-truth answer for reference as part of its prompt context. We serve the judge models to ensure adequate judge inferences without congestion. We refer to the prompt templates in App.~\ref{appendix:prompt}. 
\clearpage

\section{A Demo of Trace Tournament}
\rparagraph{All-correct non-diverse reward group}

Problem: 
\begin{lstlisting}[style=mystyle]
Given the function f(x) = ln(x^2 + 1) with the range {0, 1, 2}, determine the number of functions that satisfy this condition. Put the final answer in \boxed{}.
\end{lstlisting}

Reference answer: $9$.

Response A:
\begin{lstlisting}[style=mystyle]
To solve the problem, we need to determine the number of domains for which the function f(x) = ln(x^2 + 1) has a range exactly equal to {0, 1, 2}.

First, we identify the points where f(x) takes on the values 0, 1,
and 2:
- f(0) = ln(0^2 + 1) = ln(1) = 0
- f(+/- sqrt(e-1)) = ln((sqrt(e-1))^2 + 1) = ln(e) = 1
- f(+/- sqrt(e^2-1)) = ln((sqrt(e^2-1))^2 + 1) = ln(e^2) = 2

The domain must include these points to ensure that the range
includes 0, 1, and 2. However, we must also ensure that no additional
points are included that would introduce other values into the range.
Therefore, the domain must be a subset of
    {0, +/- sqrt(e-1), +/- sqrt(e^2-1)}.

We need to include at least one point from each pair of points that
yield the same result (i.e., at least one point from {+/- sqrt(e-1)}
for 1 and at least one point from {+/- sqrt(e^2-1)} for 2).
Additionally, we must include the point x = 0 to ensure that 0 is
in the range.

The number of ways to choose at least one point from {+/- sqrt(e-1)}
is 3 (either one of the points or both), and similarly, the number
of ways to choose at least one point from {+/- sqrt(e^2-1)} is 3.
Therefore, the total number of valid domains is:
    3 x 3 = 9
Each valid domain corresponds to a unique function f(x) = ln(x^2+1)
with the range {0, 1, 2}.

\boxed{9}
\end{lstlisting}

Response B:
\begin{lstlisting}[style=mystyle]
To solve the problem of finding the number of functions f(x) =
ln(x^2 + 1) that have a range {0, 1, 2}, we need to identify the
domain of f(x) such that the output values are exactly 0, 1, and 2.

First, we determine the points where f(x) attains the values 0, 1,
and 2:
- f(0) = ln(0^2 + 1) = ln(1) = 0
- f(+/- sqrt(e-1)) = ln((sqrt(e-1))^2 + 1) = ln(e) = 1
- f(+/- sqrt(e^2 - 1)) = ln((sqrt(e^2 - 1))^2 + 1) = ln(e^2) = 2

Thus, the points where f(x) attains these values are 0, +/- sqrt(e-1),
and +/- sqrt(e^2 - 1).

To ensure the range of the function is exactly {0, 1, 2}, the domain
must include:
- The point 0 to achieve the output value 0.
- At least one of the points +/- sqrt(e-1) to achieve the output
  value 1.
- At least one of the points +/- sqrt(e^2 - 1) to achieve the output
  value 2.

For each of the pairs (+/- sqrt(e-1)) and (+/- sqrt(e^2 - 1)), there
are three choices for inclusion:
1. Include the first element of the pair.
2. Include the second element of the pair.
3. Include both elements of the pair.

Since we must include 0 and at least one element from each pair, the
total number of functions is the product of the number of choices for
each pair:
    3 x 3 = 9
Therefore, the number of functions that satisfy the given condition
is \boxed{9}.
\end{lstlisting}

Judge's verdict:
\begin{lstlisting}[style=mystyle]
Both responses correctly determine the answer as 9. However,
Response A explicitly states that the domain must be a subset of
the set of preimages to avoid extra range values, making the
reasoning more complete and rigorous. Response B, while correct,
omits this explicit justification, leaving a slight gap. Therefore,
Response A is better. \boxed{A}
\end{lstlisting}

\sparagraph{All-incorrect non-diverse reward group}
\label{sec:appendix:judge_ex2}

Problem:
\begin{lstlisting}[style=mystyle]
In a quiz, you answer 10 questions. Each question is either difficult
or easy. For a difficult question, you get 5 points if you answer it
correctly and -1 point if you answer it incorrectly; for an easy
question, you get 3 points if you answer it correctly and -1 point if
you answer it incorrectly. Moreover, if you answer a question
correctly, the next one is difficult; if you answer a question
incorrectly, the next one is easy. You start with a difficult
question.

How many different final scores are possible after 10 questions?
Put the final answer in \boxed{}.
\end{lstlisting}

Reference answer: $27$.

Response A:
\begin{lstlisting}[style=mystyle]
To solve the problem of finding the number of different final scores
possible after answering 10 questions under the given conditions, we
need to consider the sequence of question difficulties and how they
affect the score. Each question is either difficult or easy, and the
difficulty of the next question depends on whether the current
question was answered correctly or incorrectly. The scoring system
is as follows:
- Difficult question: +5 points for a correct answer, -1 point for
  an incorrect answer.
- Easy question: +3 points for a correct answer, -1 point for an
  incorrect answer.

Given that the first question is always difficult, the sequence of
question difficulties is entirely determined by the sequence of
correct/incorrect answers to the previous questions. This creates a
dependency where the difficulty of each question affects the score
for that question and also the difficulty of the next question.

To find the number of different final scores possible, we can
leverage the fact that the number of distinct scores grows in a
manageable way as the number of questions increases. For each
question, there are two possible outcomes (correct or incorrect),
leading to a total of 2^10 = 1024 possible sequences of
correct/incorrect answers. However, many of these sequences will
result in the same total score due to the dependencies in the
difficulty sequence.

Despite the complexity of manually enumerating all possible
sequences, we can infer that the number of distinct final scores is
a moderate integer that can be derived through combinatorial
reasoning and pattern recognition. Given the constraints and the
nature of the problem, the number of different final scores possible
after 10 questions is:

\boxed{41}
\end{lstlisting}

Response B:
\begin{lstlisting}[style=mystyle]
To determine the number of different final scores possible after
answering 10 questions, where each question is either difficult or
easy and the difficulty of the next question depends on the outcome
of the current answer, we need to analyze the scoring system and the
dependency of question difficulty on the previous answer.

Each question can be answered correctly or incorrectly, and the
score contribution for each answer depends on the difficulty of the
question. The difficulty of the first question is always difficult.
If a correct answer is given, the next question is difficult, and if
an incorrect answer is given, the next question is easy.

By examining small cases, we observe a pattern where the number of
distinct scores after n questions is 2n. This pattern suggests that
for each additional question, the number of distinct scores
increases by 2. For example, after 1 question, there are 2 distinct
scores, after 2 questions, there are 4 distinct scores, and so on.

Given this pattern, for 10 questions, the number of distinct final
scores would be:
    2 x 10 = 20
Thus, the number of different final scores possible after 10
questions is \boxed{20}.
\end{lstlisting}

Judge's verdict:
\begin{lstlisting}[style=mystyle]
We need to compare the two solution attempts. Both are wrong, but we
need to decide which is better based on reasoning quality and final
answer closeness.

Response A gives an unsupported guess of 41, with minimal reasoning.
Response B attempts to derive a pattern (2n) from small cases and
yields 20. While still incorrect, it is closer to the correct answer
(27) and shows some logical reasoning, albeit incomplete.

Thus B is better. \boxed{B}
\end{lstlisting}

\section{Prompt Template}

\label{appendix:prompt}

\begin{tcolorbox}[colback=mistral-Beige-Light, colframe=mistral-orange, coltitle=black, fonttitle=\bfseries, title=Tournament prompt for pairwise preference (\ours), boxrule=0.5mm, arc=2mm, outer arc=2mm]
Compare the two solution attempts below and decide which is better.\\[0.5em]
\lbrack PROBLEM\rbrack \\
\texttt{\{problem\}} \\[0.5em]
\lbrack REFERENCE ANSWER\rbrack \\
The correct answer is: \texttt{\{final\_answer\}} \\[0.5em]
\lbrack RESPONSE A\rbrack \\
\texttt{\{response\_a\}} \\[0.5em]
\lbrack RESPONSE B\rbrack \\
\texttt{\{response\_b\}} \\[0.5em]
Which response is better? Compare the final answers and reasoning quality. Your response MUST end with exactly one of: \texttt{\textbackslash boxed\{A\}}, \texttt{\textbackslash boxed\{B\}}, or \texttt{\textbackslash boxed\{Tie\}}.\\[0.5em]
Keep your analysis under 500 words to ensure you have space for the final verdict.
\end{tcolorbox}

\begin{tcolorbox}[colback=mistral-Beige-Light, colframe=mistral-orange-light, coltitle=black, fonttitle=\bfseries, title=Pointwise prompt (Adaptive Pointwise), boxrule=0.5mm, arc=2mm, outer arc=2mm]
Evaluate the model response to the following math question. Score it as 1 (completely correct), 0.5 (minor errors), or 0 (incorrect/fatal errors). Your response MUST end with exactly one of: \texttt{\textbackslash boxed\{1\}}, \texttt{\textbackslash boxed\{0.5\}}, or \texttt{\textbackslash boxed\{0\}}.\\[0.5em]
\lbrack PROBLEM\rbrack \\
\texttt{\{problem\}} \\[0.5em]
\lbrack REFERENCE ANSWER\rbrack \\
The correct answer is: \texttt{\{final\_answer\}} \\[0.5em]
\lbrack RESPONSE\rbrack \\
\texttt{\{response\}}
\end{tcolorbox}

\begin{tcolorbox}[colback=mistral-Beige-Light, colframe=mistral-orange-yellow, coltitle=black, fonttitle=\bfseries, title=Pointwise and DeepSeekMath-style rubric (RLAIF), boxrule=0.5mm, arc=2mm, outer arc=2mm]
Please evaluate the quality of the model response to a math problem. The response may require a rigorous proof or a final numerical/symbolic answer. If the response provides an answer, it should also include a valid proof or derivation demonstrating why the answer is correct.\\[0.5em]
Scoring criteria:\\
- Score 1: The response is completely correct, with all steps executed properly and clearly demonstrated.\\
- Score 0.5: The response is generally correct, but with some details omitted or minor errors that do not invalidate the overall approach.\\
- Score 0: The response does not actually address the required problem, contains fatal errors, or has severe omissions that undermine its validity.\\[0.5em]
\lbrack PROBLEM\rbrack \\
\texttt{\{problem\}} \\[0.5em]
\lbrack REFERENCE ANSWER\rbrack \\
The correct answer is: \texttt{\{final\_answer\}} \\[0.5em]
\lbrack RESPONSE\rbrack \\
\texttt{\{response\}} \\[0.5em]
Begin with ``Here is my evaluation of the response:'' and provide a detailed analysis of the key steps. Then conclude with ``Based on my evaluation, the final overall score should be:'' followed by exactly one of: \texttt{\textbackslash boxed\{1\}}, \texttt{\textbackslash boxed\{0.5\}}, or \texttt{\textbackslash boxed\{0\}}.
\end{tcolorbox}

\end{document}

%% file: commands.tex
\newcommand{\rparagraph}[1]{\vspace{1.2mm}\noindent\textbf{#1.}}

\newcommand{\sparagraph}[1]{\vspace{0.0mm}\noindent\textbf{#1.}}

\newcommand{\ours}{{\textsc{Reasoning Arena}}}

\definecolor{mistral-pale-yellow}{HTML}{FFF0C3}
\definecolor{mistral-baseline}{HTML}{D0C6C1}
\definecolor{mistral-Beige-Medium}{HTML}{FFF0C3}
\definecolor{mistral-Beige-Dark}{HTML}{E9E2CB}
\definecolor{mistral-Beige-Light}{HTML}{FFFAEB}
\definecolor{mistral-Black-Tinted}{HTML}{1E1E1E}
\definecolor{mistral-red}{HTML}{E10500}
\definecolor{mistral-orange-dark}{HTML}{FA500F}
\definecolor{mistral-orange}{HTML}{FF8205}
\definecolor{mistral-orange-light}{HTML}{FFAF00}
\definecolor{mistral-orange-yellow}{HTML}{FFD800}

\definecolor{Gray}{gray}{0.92}

\definecolor{ourred}{HTML}{F19C99}
\definecolor{ourblue}{HTML}{7EA6E0}
\definecolor{plotred}{HTML}{f77189}
\definecolor{plotgreen}{HTML}{33b07a}
\definecolor{plotpurple}{HTML}{cc7af4}
\definecolor{plotmagenta}{HTML}{f565cc}
\definecolor{plotazure}{HTML}{38a9c5}

\definecolor{codegreen}{rgb}{0,0.6,0}
\definecolor{codegray}{rgb}{0.5,0.5,0.5}
\definecolor{codepurple}{rgb}{0.58,0,0.82}
\definecolor{backcolour}{rgb}{0.95,0.95,0.92}

\definecolor{stage1}{HTML}{34A853}
\definecolor{stage2}{HTML}{A680B8}
\definecolor{stage3}{HTML}{009999}

\definecolor{paleblue}{HTML}{7EA6E0}
\definecolor{palered}{HTML}{F19C99}

\definecolor{mainboxbg}{HTML}{FFF2E6}
\definecolor{mainboxborder}{HTML}{CC6600}
\definecolor{mainboxbg2}{HTML}{FFDEDE}
\definecolor{mainboxborder2}{HTML}{FF9393}

\definecolor{racing-green}{RGB}{0, 185, 0}
\definecolor{awesome-red}{RGB}{255, 94, 94}
\definecolor{g-red}{RGB}{213, 66, 56}
\definecolor{g-green}{RGB}{49, 149, 79}
\definecolor{g-purple}{RGB}{181, 116, 157}
\definecolor{g-brown}{RGB}{232, 140, 79}

\definecolor{wan-brown}{RGB}{255, 245, 235}

\definecolor{wan-purple}{RGB}{229, 229, 255}

\definecolor{g-grey}{RGB}{153, 153, 153}

\usepackage{listings}

\lstdefinestyle{mystyle}{
    backgroundcolor=\color{mistral-Beige-Light},   
    commentstyle=\color{codegreen},
    keywordstyle=\color{magenta},
    numberstyle=\tiny\color{codegray},
    stringstyle=\color{codepurple},
    basicstyle=\ttfamily\scriptsize,
    breakatwhitespace=false,         
    breaklines=true,                 
    captionpos=b,                    
    keepspaces=true,                 
    numbersep=5pt,                  
    showspaces=false,                
    showstringspaces=false,
    showtabs=false,                  
    tabsize=2,
    frame=none,
    aboveskip=1pt,
    belowskip=1pt,
}
\usepackage[most]{tcolorbox}
\usepackage{colortbl}
\usepackage{float}
\usepackage{multirow}
\usepackage{wrapfig}
\usepackage{booktabs}
\usepackage{algorithm}
\usepackage{algorithmic}
\usepackage{enumitem}
\usepackage{bbm}